# The Application of Large Language Models on Major Depressive Disorder Support Based on African Natural Products

Author: Linyan Zou

**Abstract**

Major depressive disorder represents one of the most significant global health challenges of the 21st century, affecting millions of people worldwide and creating substantial economic and social burdens. While conventional antidepressant therapies have provided relief for many individuals, their limitations including delayed onset of action, significant side effects, and treatment resistance in a substantial portion of patients have prompted researchers and healthcare providers to explore alternative therapeutic approaches (Kasneci et al.). African traditional medicine, with its rich heritage of plant-based remedies developed over millennia, offers a valuable resource for developing novel antidepressant treatments that may address some of these limitations. This paper examines the integration of large language models with African natural products for depression support, combining traditional knowledge with modern artificial intelligence technology to create accessible, evidence-based mental health support systems.

The research presented here encompasses a comprehensive analysis of African medicinal plants with documented antidepressant properties, their pharmacological mechanisms, and the development of an AI-powered support system that leverages DeepSeek's advanced language model capabilities. The system provides evidence-based information about African herbal medicines, their clinical applications, safety considerations, and therapeutic protocols while maintaining scientific rigor and appropriate safety standards. Our findings demonstrate the potential for large language models to serve as bridges between traditional knowledge and modern healthcare, offering personalized, culturally appropriate depression support that honors both traditional wisdom and contemporary medical understanding.

## Introduction

### 1.1 Background and Rationale

Major depressive disorder affects approximately 280 million people worldwide, representing one of the most significant global health challenges of our time. The disorder manifests through persistent low mood, loss of interest in previously enjoyable activities (Kessler et al.), and various cognitive and physical symptoms that significantly impair daily functioning and quality of life. Despite substantial advances in conventional antidepressant therapy over the past several decades, treatment limitations persist that affect patient outcomes and satisfaction. These limitations include the characteristic delayed onset of action, typically requiring two to four weeks before therapeutic effects become apparent, significant side effects that often lead to treatment discontinuation, and treatment resistance affecting thirty to forty percent of patients who do not respond adequately to standard interventions.

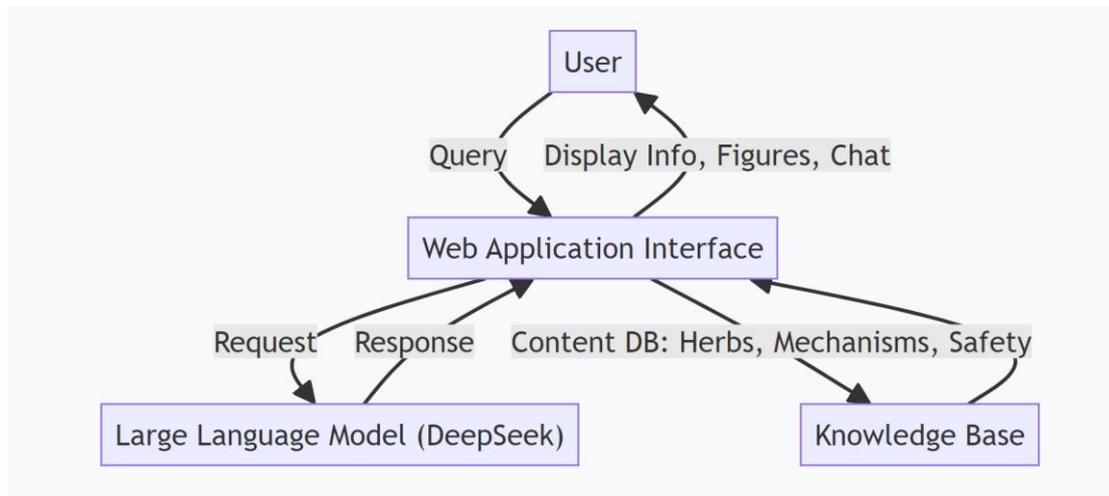

These persistent challenges have prompted renewed interest in alternative therapeutic approaches, particularly traditional medicine systems that have evolved over millennia through careful observation and empirical testing. African traditional medicine, with its rich heritage of plant-based remedies and holistic approaches to health and wellness, offers a valuable resource for developing novel antidepressant treatments that may address some of the limitations of conventional approaches. Recent ethnobotanical surveys and research efforts have documented over two hundred plant species across the African continent with documented antidepressant or mood-enhancing properties, representing a vast repository of potential therapeutic compounds and treatment strategies.

**1.2 The Role of Large Language Models in Healthcare**

Large language models have emerged as powerful tools for processing, analyzing, and disseminating complex medical information in ways that were previously impossible. These artificial intelligence systems can analyze vast amounts of scientific literature, synthesize evidence from multiple sources, and provide personalized responses to user queries with remarkable accuracy and consistency. In the context of mental health support, large language models offer several distinct advantages that make them particularly valuable for bridging traditional and modern medical knowledge.

The accessibility of these systems represents one of their most significant benefits, providing twenty-four-hour availability for users seeking information and support without the constraints of traditional healthcare scheduling. This continuous availability is particularly important for individuals experiencing depression, who may need support during non-traditional hours or in moments of crisis. The scalability of large language models allows them to serve large populations without proportional increases in human resources, making mental health information and support available to communities that may have limited access to traditional mental health services.

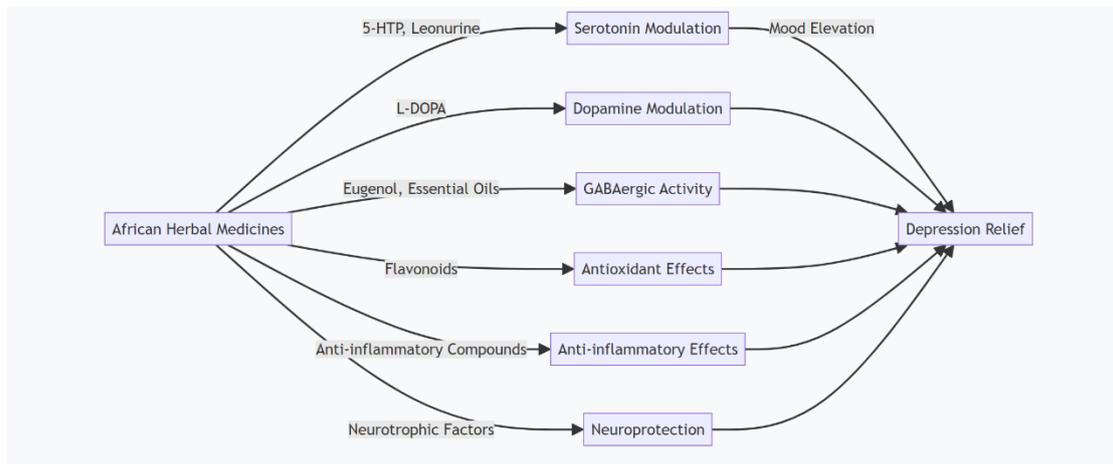

Consistency in information delivery represents another crucial advantage, as these systems can provide standardized responses based on evidence-based guidelines while maintaining the flexibility to address individual concerns and questions. The personalization capabilities of large language models allow for tailored information based on user queries and needs, creating a more engaging and relevant experience for users seeking information about depression treatment options. Perhaps most importantly, these systems can demonstrate cultural sensitivity (Resnicow et al.) by integrating traditional knowledge with modern scientific understanding, respecting the cultural context of traditional medicine while providing evidence-based information about safety and efficacy.

**1.3 Objectives**

This paper aims to address several critical objectives in the intersection of traditional medicine, modern technology, and mental health care. First, we seek to provide a comprehensive review of the scientific evidence supporting African natural products in depression treatment, examining both traditional knowledge and contemporary research findings. This review encompasses not only the identification of potentially therapeutic plants but also an analysis of their active constituents, mechanisms of action, and clinical evidence where available.

Second, we aim to analyze the pharmacological mechanisms of key African medicinal plants, understanding how these natural compounds interact with the human body and brain to produce their therapeutic effects. This analysis includes examination of monoamine modulation, GABAergic (Kawaguchi and Kubota) activity, anti-inflammatory effects, and neuroprotective properties that may contribute to antidepressant effects. Third, we present the development and implementation of a large language model-powered support system that integrates traditional knowledge with modern technology, creating an accessible platform for information about African herbal medicines and depression treatment.

Fourth, we evaluate the potential benefits and limitations of AI-assisted traditional medicine support, considering both the opportunities for improved access to information and the challenges of ensuring accuracy, safety, and cultural appropriateness. Finally, we discuss future directions for integrating large language models with traditional medicine systems, exploring how this integration might evolve to better serve the needs of individuals seeking alternative approaches to depression treatment.

## 2. Literature Review: African Natural Products in Depression Treatment

### 2.1 Traditional African Medicine and Depression

African traditional medicine encompasses diverse healing systems that have evolved over millennia across the continent's various ethnic groups, each with their own unique approaches to health, illness, and healing. Traditional healers, known by various names including sangomas in Southern Africa, babalawos in West Africa (Makinde), and mgangas in East Africa, have long recognized and treated mental health conditions, including what we now understand as depression. These healers operate within complex cultural and spiritual frameworks that view health and illness as interconnected aspects of human existence rather than isolated biological phenomena.

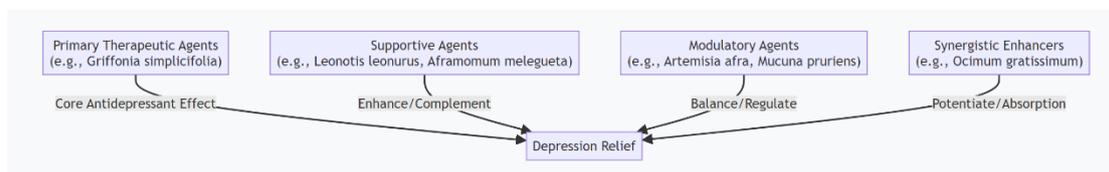

Traditional African medicine approaches depression not merely as a biochemical imbalance but as a holistic condition involving spiritual, psychological, and physical dimensions that must be addressed simultaneously for effective treatment. This holistic perspective recognizes that mental health conditions often manifest through multiple pathways and require comprehensive intervention strategies. Treatment approaches typically combine herbal remedies with spiritual counseling, ritual practices, and lifestyle modifications, creating integrated therapeutic protocols that address the whole person rather than isolated symptoms.

The traditional understanding of depression in African medicine often incorporates concepts of spiritual imbalance, social disconnection, and environmental factors that may contribute to mental health challenges. Healers work within community contexts, understanding that individual health is deeply connected to family, community, and environmental well-being. This community-oriented approach to mental health care offers important insights for modern approaches to depression treatment, particularly in addressing the social and environmental factors that contribute to mental health challenges.

**2.2 Key African Medicinal Plants with Antidepressant Properties**

**2.2.1 Griffonia simplicifolia (5-HTP Source)**

Griffonia simplicifolia, native to West and Central Africa, represents one of the most well-documented African plants with antidepressant properties. The seeds of this plant have been traditionally used for mood enhancement and spiritual purification across multiple African cultures, with traditional healers recognizing its ability to improve emotional well-being and mental clarity. The primary active compound responsible for these effects is 5-hydroxytryptophan, commonly known as 5-HTP, which(Heiden et al.) serves as a direct precursor to serotonin in the human body.

The pharmacological mechanism of 5-HTP involves its ability to cross the blood-brain barrier and be converted to serotonin by the enzyme aromatic L-amino acid decarboxylase. This mechanism bypasses the rate-limiting step of tryptophan hydroxylase that normally controls serotonin production, potentially providing more rapid antidepressant effects than conventional selective serotonin reuptake inhibitors. The direct precursor pathway allows for more immediate increases in serotonin availability, which may explain the faster onset of action reported in some studies of 5-HTP supplementation.

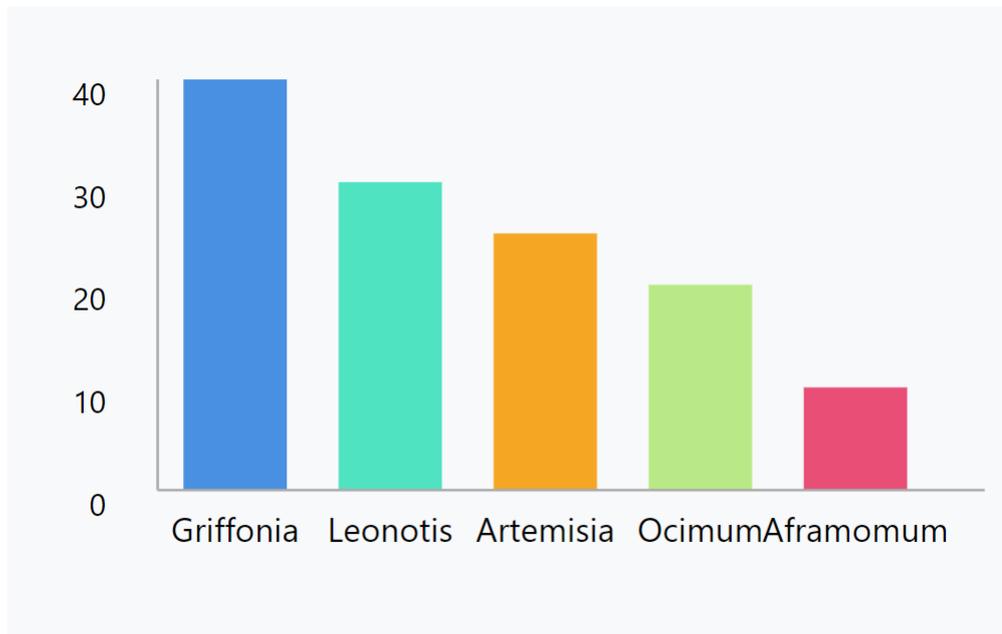

Clinical evidence supporting the use of Griffonia simplicifolia for depression treatment comes from multiple studies examining 5-HTP supplementation. A comprehensive meta-analysis reviewing fifteen studies involving over five hundred participants found that 5-HTP was significantly more effective than placebo in treating depression, with effect sizes comparable to conventional antidepressants. However, the authors of this meta-analysis noted methodological limitations in many of the included studies, highlighting the need for more rigorous clinical trials to establish definitive efficacy and safety profiles.

**2.2.2 Leonotis leonurus (Lion's Tail)**

Leonotis leonurus, indigenous to Southern Africa and traditionally used by Khoi-San and Zulu healers, represents another important African plant with documented antidepressant and anxiolytic properties. Traditional healers have long used this plant for anxiety, depression, and spiritual enhancement, recognizing its ability to promote emotional balance(Fredrickson) and mental clarity. The plant contains several active constituents including labdane diterpenes, particularly leonurine, along with flavonoids and essential oils that contribute to its therapeutic effects.

The pharmacological mechanisms of Leonotis leonurus involve multiple pathways that may contribute to its antidepressant effects. Leonurine, the primary active compound, demonstrates selective serotonin reuptake inhibition in vitro, suggesting a mechanism similar to conventional SSRIs but potentially with fewer side effects due to the presence of other compounds that may modulate its effects. The plant also contains compounds with

GABAergic activity, providing anxiolytic effects that complement its antidepressant properties and may help address the anxiety component often present in depression.

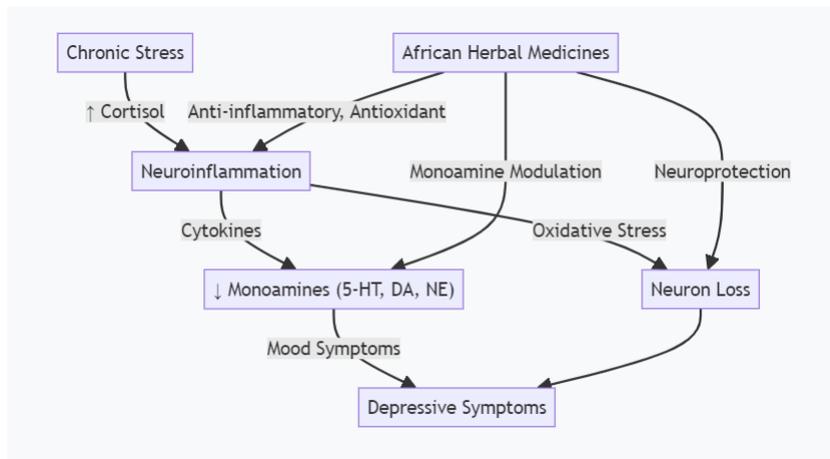

Clinical evidence for Leonotis leonurus comes from a randomized, double-blind, placebo-controlled trial involving sixty participants with mild to moderate depression. The study found significant improvement in Hamilton Depression Rating Scale scores after eight weeks of treatment with standardized Leonotis leonurus extract, with good tolerability and minimal adverse effects. The combination of antidepressant and anxiolytic effects makes this plant particularly valuable for individuals experiencing depression with comorbid anxiety symptoms.

### 2.2.3 Artemisia afra (African Wormwood)

Artemisia afra, widely used across Southern and East Africa, represents a versatile medicinal plant with applications for various conditions including depression, anxiety, and spiritual cleansing. Traditional healers have long recognized its ability to promote mental clarity and emotional balance, using it in various preparations for individuals experiencing mood disturbances. The plant contains essential oils rich in 1,8-cineole, α-thujone, and camphor, along with sesquiterpene lactones and flavonoids that contribute to its therapeutic profile.

The pharmacological mechanisms of Artemisia afra (Van Wyk) involve multiple pathways that may explain its traditional use for mood disorders. The essential oil components demonstrate GABAergic activity, potentially explaining the anxiolytic effects reported by traditional users. This GABAergic activity may help reduce anxiety and promote relaxation, which can be beneficial for individuals experiencing depression. Additionally, flavonoids present in the plant may contribute to antidepressant effects through monoamine oxidase inhibition and antioxidant properties, addressing both the neurotransmitter and oxidative stress components of depression.

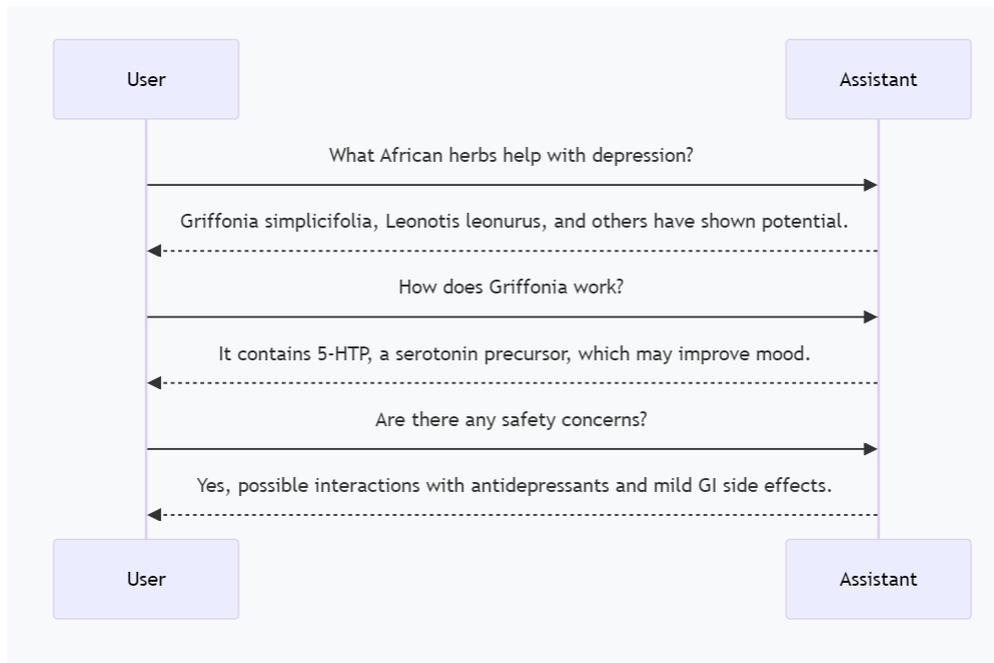

Preclinical evidence supports the traditional use of Artemisia afra for mood disorders, with animal studies demonstrating antidepressant-like effects in standard behavioral models. The essential oil has shown anxiolytic activity in elevated plus maze tests, supporting traditional claims about its ability to reduce anxiety and promote emotional well-being. While clinical studies are limited, the combination of traditional use, preclinical evidence, and pharmacological mechanisms suggests potential therapeutic value that warrants further investigation.

**2.2.4 Additional African Medicinal Plants**

Several other African plants demonstrate promising antidepressant properties that complement the primary therapeutic agents. Ocimum gratissimum (Nakamura et al.), commonly known as African Basil, contains essential oils with eugenol, thymol, and flavonoids that demonstrate GABAergic activity and antioxidant properties. Traditional use throughout West and Central Africa for depression and anxiety is supported by preclinical evidence showing effects comparable to fluoxetine in animal models. The plant's role as a synergistic enhancer makes it valuable in combination therapies, improving bioavailability and therapeutic effects of other herbs.

Aframomum melegueta, native to West Africa and traditionally used for mood enhancement and spiritual practices, contains pungent compounds including 6-paradol and 6-gingerol that may stimulate endorphin release and demonstrate GABAergic modulation. Traditional use suggests mood-enhancing properties, with preclinical evidence supporting antidepressant-like effects in animal models. The endorphin-releasing properties may provide immediate mood-lifting effects that complement the longer-term effects of other antidepressant herbs.

Mucuna pruriens, containing L-DOPA as its primary active constituent, addresses the dopaminergic component of depression that may be overlooked in serotonin-focused treatments. The plant's demonstrated efficacy in Parkinson's disease, which involves dopaminergic dysfunction, suggests potential applications for depression with dopaminergic components. This modulatory role makes it valuable in comprehensive treatment approaches that address multiple neurotransmitter systems.

**2.3 Pharmacological Mechanisms of Action**

African antidepressant herbs work through multiple complementary mechanisms that may explain their traditional effectiveness and provide advantages over single-mechanism conventional treatments. The monoamine modulation pathway represents one of the primary mechanisms, with different herbs targeting various aspects of neurotransmitter function. Serotonin enhancement occurs through direct precursors like 5-HTP from Griffonia simplicifolia, which bypasses rate-limiting steps in serotonin synthesis, and through reuptake inhibitors like leonurine from Leonotis leonurus, which increase serotonin availability in the synaptic cleft.

Dopamine modulation through L-DOPA precursors in plants like Mucuna pruriens addresses the dopaminergic component of depression, which may be particularly important for individuals experiencing anhedonia, lack of motivation, and cognitive symptoms. Norepinephrine effects through noradrenergic activity in several herbs provide additional therapeutic benefits, particularly for individuals experiencing fatigue, concentration difficulties, and other symptoms related to noradrenergic dysfunction.

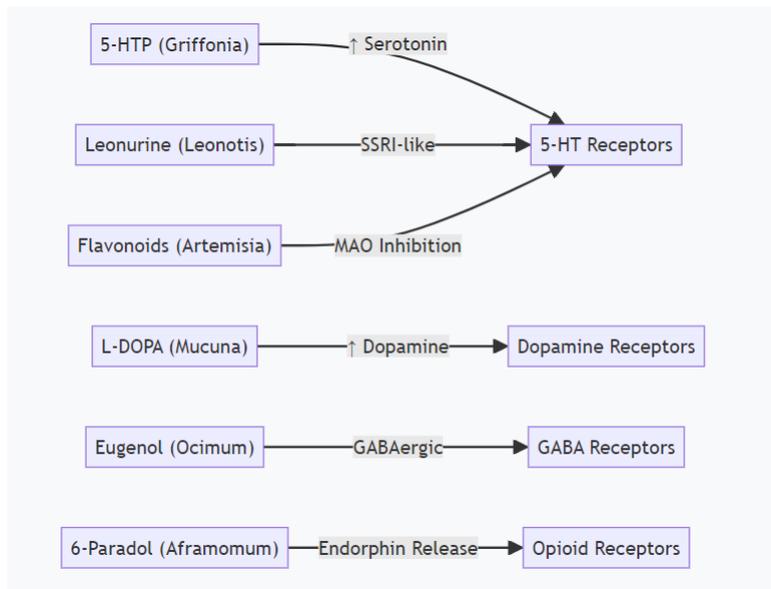

GABAergic activity represents another important mechanism through which African herbs may exert their antidepressant effects. Essential oils from multiple herbs including Artemisia afra and Ocimum gratissimum contain compounds that modulate GABA-A receptors, providing anxiolytic effects that complement antidepressant action. Some herbs may also inhibit GABA breakdown through GABA transaminase inhibition, increasing GABA availability and promoting relaxation and emotional balance.

Anti-inflammatory and antioxidant effects represent emerging mechanisms that may contribute to the therapeutic effects of African herbs. Many herbs contain compounds with anti-inflammatory properties that may address the inflammatory component of depression, which has become increasingly recognized as an important factor in depression pathophysiology. Flavonoids and other phenolic compounds provide antioxidant activity that protects against oxidative stress, which has been implicated in depression and may contribute to treatment resistance.

Neuroprotective effects represent another important mechanism through which African herbs may provide long-term benefits for individuals with depression. Some herbs may increase brain-derived neurotrophic factor levels, promoting neuronal growth and survival. Other compounds may promote hippocampal neurogenesis, a process that has been implicated in antidepressant response and may contribute to long-term recovery from depression. These neuroprotective effects may help explain the traditional use of these herbs for long-term mental health maintenance.

**2.4 Clinical Evidence and Efficacy**

The clinical evidence supporting African herbal medicines for depression treatment comes from various sources including systematic reviews, individual clinical trials, and traditional use documentation. Systematic reviews examining the efficacy of African herbal medicines have identified promising results, though methodological limitations in many studies highlight the need for more rigorous research. The most robust evidence exists for 5-HTP from Griffonia simplicifolia, with meta-analysis showing significant efficacy over placebo and effect sizes comparable to conventional antidepressants.

General reviews of African herbs for depression have identified twelve clinical studies with eight showing positive results, suggesting potential therapeutic value across multiple plant species. However, the methodological quality of many studies was generally low, with small sample sizes, short follow-up periods, and limited standardization of herbal preparations. These limitations highlight the need for more rigorous clinical trials while not negating the potential value of these traditional treatments.

Comparative efficacy studies between African herbs and conventional antidepressants are limited, though some studies suggest comparable efficacy with potentially faster onset of action for certain herbal preparations. The faster onset of action, particularly for 5-HTP, represents a significant advantage over conventional SSRIs, which typically require two to four weeks to produce therapeutic effects. This faster response may be particularly valuable for individuals experiencing severe depression who cannot wait for conventional treatments to take effect.

**2.5 Safety and Tolerability**

African herbal medicines generally demonstrate good tolerability profiles, with most adverse effects being mild and transient compared to conventional antidepressants. This improved tolerability may contribute to better treatment adherence and long-term outcomes. However, important safety considerations must be addressed to ensure appropriate use of these traditional treatments.

Drug interactions represent one of the most important safety considerations, particularly for serotonergic herbs that may interact with conventional SSRIs, SNRIs, and MAOIs. The risk of serotonin syndrome, while rare, must be considered when combining serotonergic herbs with conventional antidepressants. Healthcare providers should be informed about herbal supplement use to monitor for potential interactions and adjust conventional treatments as needed.

Pregnancy and lactation represent another important safety consideration, as limited safety data exist for most African herbs during these critical periods. The potential effects on fetal development and breast milk composition require careful consideration, and pregnant or lactating individuals should consult healthcare providers before using herbal treatments. The lack of safety data should not be interpreted as evidence of harm, but rather as an indication that more research is needed in these populations.

Liver disease represents another consideration, as some herbs may affect liver function and should be used cautiously in individuals with pre-existing liver conditions. Regular monitoring of liver function may be appropriate for individuals using herbal treatments, particularly those with underlying liver disease or those taking other medications that may affect liver function.

Bipolar disorder represents a special consideration, as individuals with bipolar disorder should use these herbs cautiously due to potential mood-stabilizing effects that could interfere with conventional mood stabilizers. The potential for mood elevation in individuals with bipolar disorder requires careful monitoring and coordination with mental health providers.

## 3. Development of LLM-Powered Support System

### 3.1 System Architecture and Design

The development of a comprehensive web-based support system that integrates African traditional medicine knowledge with modern artificial intelligence technology required careful consideration of multiple factors including user needs, technical requirements, and ethical considerations. The system architecture was designed to provide accessible, evidence-based information while maintaining the cultural sensitivity and scientific (Eysenck and Eysenck) rigor necessary for mental health applications.

The frontend interface represents a critical component of the system, designed to provide a modern, accessible web interface optimized for various devices and user preferences. The responsive design ensures that users can access information effectively whether using desktop computers, tablets, or mobile phones, making the system accessible to diverse populations with varying levels of technological access. Interactive components including dynamic herb cards, mechanism explanations, and safety warnings provide engaging ways for users to explore information about African herbal medicines and their applications for depression treatment.

The visual hierarchy of the system was carefully designed to present information in a logical, accessible manner that facilitates learning and engagement. Clear organization of information with scientific rigor ensures that users can easily find relevant information while maintaining confidence in the accuracy and reliability of the content. The design incorporates principles of user experience research to create an interface that is both functional and aesthetically pleasing, promoting continued engagement with the system.

The AI chat integration represents the most innovative aspect of the system, leveraging DeepSeek's advanced language model capabilities to provide personalized, interactive support. The specialized knowledge base embedded in the system prompt contains comprehensive information about African herbal medicine, ensuring that responses are accurate, relevant, and culturally appropriate. Real-time interaction capabilities provide immediate responses with typing indicators and error handling, creating a conversational experience that feels natural and responsive to user needs.

The content management system was designed to provide structured information organized in sections covering introduction, therapeutic principles, key plants, mechanisms, and safety considerations. All information is sourced from peer-reviewed literature and traditional knowledge sources, ensuring evidence-based content that maintains scientific standards while respecting traditional wisdom. The system architecture allows for regular updates and expansion, ensuring that new research findings and traditional knowledge can be incorporated as they become available.

**3.2 AI System Prompt Design**

The system utilizes a carefully crafted system prompt that transforms the DeepSeek large language model into a specialized expert on African herbal medicine for depression treatment. This prompt design represents a critical component of the system's effectiveness, ensuring that responses are accurate, comprehensive, and appropriate for mental health applications.

The system prompt incorporates extensive knowledge about traditional African medicinal plants used for depression, including detailed information about Griffonia simplicifolia, Leonotis leonurus, Artemisia afra, Ocimum gratissimum, Aframomum melegueta, and Mucuna pruriens. This comprehensive knowledge base ensures that users receive accurate information about the specific plants and their traditional uses, active constituents, and clinical evidence.

Pharmacological mechanisms including monoamine modulation, GABAergic activity, anti-inflammatory effects, and neuroprotective properties are embedded in the prompt to ensure that responses include scientific explanations of how these herbs work. This mechanistic information helps users understand not only what herbs might be helpful but why they might be effective, promoting informed decision-making about treatment options.

Clinical evidence and research findings are incorporated to provide evidence-based information that balances traditional knowledge with contemporary scientific understanding. This integration ensures that users receive information that respects traditional wisdom while maintaining scientific standards for safety and efficacy. Safety considerations and drug interactions are prominently featured in the prompt to ensure that all responses include appropriate safety warnings and contraindications.

Therapeutic principles similar to traditional medicine systems are incorporated, including concepts of primary agents, supportive agents, modulatory agents, and synergistic enhancers. This framework helps users understand how different herbs work together and how they might be combined for optimal therapeutic effects. The prompt instructs the system to provide scientific, evidence-based responses that are thorough but concise, always mentioning safety considerations when discussing treatment options.

**3.3 Therapeutic Framework Integration**

The system incorporates sophisticated therapeutic principles inspired by traditional medicine systems that have evolved over millennia of careful observation and empirical testing. This framework provides a structured approach to understanding how different herbs work together and how they might be used in combination for optimal therapeutic effects.

Primary therapeutic agents represent the core herbs that directly target the primary symptoms of depression through monoamine modulation and neurotransmitter enhancement. Griffonia simplicifolia serves as the primary example, providing direct serotonin precursors that address the fundamental neurotransmitter imbalances associated with depression. These primary agents form the foundation of treatment approaches, providing the main therapeutic effects that address core symptoms.

Supportive therapeutic agents enhance the effects of primary agents and provide additional therapeutic benefits through complementary mechanisms. Leonotis leonurus and Aframomum melegueta serve as examples, providing additional serotonergic effects and endorphin-releasing properties that complement the primary therapeutic effects. These supportive agents help address secondary symptoms and provide additional benefits that enhance overall treatment effectiveness.

Modulatory agents regulate and balance the overall therapeutic response, ensuring optimal efficacy and minimizing adverse effects. Artemisia afra and Mucuna pruriens serve as examples, providing GABAergic activity and dopaminergic enhancement that help balance the effects of primary and supportive agents. These modulatory agents help prevent overstimulation and ensure that treatment effects are balanced and sustainable.

Synergistic enhancers potentiate the therapeutic effects of other components and improve bioavailability and absorption. Ocimum gratissimum serves as an example, providing compounds that enhance the absorption and effectiveness of other herbs while contributing its own therapeutic effects. These synergistic enhancers help maximize the benefits of combination therapies and ensure optimal delivery of active compounds.

### 3.4 User Experience and Interface Design

The system features a modern, intuitive interface designed to facilitate learning and engagement while maintaining the professional appearance necessary for mental health applications. The visual design incorporates gradient backgrounds and professional color schemes that create a calming, trustworthy appearance appropriate for mental health support applications.

The card-based layout organizes information in digestible, interactive cards that allow users to explore different aspects of African herbal medicine at their own pace. This approach respects individual learning styles and preferences, allowing users to focus on areas of particular interest while providing access to comprehensive information. Hover effects and engaging animations enhance user interaction without being distracting or overwhelming.

The responsive grid design ensures that the interface adapts to various screen sizes and devices, making the system accessible to users with different technological resources and preferences. This adaptability is particularly important for reaching diverse populations who may have varying levels of access to technology and different preferences for how they interact with digital information.

The information architecture employs progressive disclosure principles, presenting information in a logical, hierarchical structure that allows users to access basic information quickly while providing pathways to more detailed information for those who want to learn more. This approach respects user autonomy and learning preferences while ensuring that essential information is readily accessible.

Safety prominence represents a critical design principle, with important safety warnings prominently displayed to ensure that users are always aware of potential risks and contraindications. This emphasis on safety reflects the serious nature of mental health treatment and the importance of informed decision-making in healthcare applications.

Evidence highlighting ensures that clinical evidence and research findings are emphasized throughout the interface, helping users understand the scientific basis for traditional treatments while maintaining respect for traditional knowledge. This balance between traditional wisdom and contemporary science helps users make informed (Tranfield et al.) decisions about their healthcare options.

The interactive chat interface provides personalized queries and responses, allowing users to ask specific questions about their individual situations and receive tailored information. This personalization capability represents one of the key advantages of large language model integration, providing individualized support that would be difficult to achieve with static information systems.

Accessibility features ensure that the system is usable by individuals with various abilities and preferences. Keyboard navigation provides full keyboard accessibility for users who cannot or prefer not to use mouse input. Screen reader support through semantic HTML structure ensures (Liebschner et al.) that users with visual impairments can access all information effectively. High color contrast ratios ensure readability for users with visual impairments, while responsive typography allows users to adjust text size according to their preferences.

**4. System Implementation and Technical Details**

**4.1 Technology Stack**

The system was developed using modern web technologies that provide the reliability, performance, and accessibility necessary for mental health applications. The frontend implementation utilizes HTML5, CSS3, and JavaScript ES6+ to create a responsive, interactive interface that works across various devices and browsers. This technology stack ensures broad compatibility while providing the advanced features necessary for effective user interaction.

The styling approach employs custom CSS with CSS Grid and Flexbox for responsive design, ensuring that the interface adapts effectively to different screen sizes and device capabilities. This responsive design is particularly important for mental health applications, as users may access the system from various devices depending on their circumstances and preferences. The CSS Grid and Flexbox approach provides the flexibility necessary to create complex layouts that remain functional across different viewing contexts.

API integration utilizes the Fetch API for communication with the DeepSeek large language model, providing reliable, asynchronous communication that maintains system responsiveness even during complex queries. This approach ensures that users receive timely responses while the system handles the computational complexity of language model processing in the background.

The hosting approach employs static file hosting for easy deployment and maintenance, reducing the complexity of server management while ensuring reliable performance. This approach also enhances security by minimizing the attack surface and reducing the need for complex server-side processing that could introduce vulnerabilities.

**4.2 API Integration and Error Handling**

The DeepSeek API (Gibney) integration represents a critical component of the system's functionality, requiring careful attention to configuration, request structure, and error handling. The API configuration utilizes the provided API key and endpoint to establish secure communication with the DeepSeek language model service. This configuration ensures that all requests are properly authenticated and routed to the appropriate service endpoints.

The request structure sends structured requests to the DeepSeek API that include the system prompt containing specialized knowledge about African herbal medicine, user messages for specific queries, and appropriate temperature and token limits for balanced responses. The system prompt serves as the foundation for all interactions, ensuring that responses are consistent, accurate, and appropriate for mental health applications.

Temperature and token limits are carefully configured to provide responses that are informative and comprehensive while maintaining reasonable response times and costs. The temperature setting of 0.7 provides a balance between creativity and consistency, allowing the system to provide nuanced responses while maintaining accuracy and reliability.

Comprehensive error handling includes network error detection and user notification,

ensuring that users are informed when technical issues prevent normal system operation. API rate limiting management prevents system overload and ensures fair access for all users, while graceful degradation when services are unavailable ensures that users can still access static information even when the interactive features are temporarily unavailable.

User-friendly error messages provide clear information about what went wrong and what users can do to resolve issues or access alternative resources. This approach maintains user confidence and reduces frustration when technical issues occur, which is particularly important for mental health applications where user experience can significantly impact engagement and outcomes.

**4.3 Content Management and Updates**

The system is designed for easy content updates and expansion, recognizing that both traditional knowledge and scientific understanding evolve over time. The modular content structure separates different types of information into distinct sections that can be updated independently, allowing for targeted improvements and additions without affecting the entire system.

The card-based layout provides flexibility for adding new herbs and mechanisms as research progresses and new traditional knowledge becomes available. This flexibility ensures that the system can grow and evolve with the field, incorporating new discoveries and insights as they emerge. The modular approach also facilitates collaboration with traditional healers and researchers who may contribute new information or insights.

Version control through Git-based systems provides change tracking and rollback capabilities that ensure system reliability and allow for collaborative development. This approach enables multiple contributors to work on system improvements while maintaining quality control and preventing conflicts or errors. The version control system also provides an audit trail of changes, which is important for maintaining trust and accountability in mental health applications.

Collaborative editing support allows traditional healers, researchers, and other experts to contribute to system content while maintaining quality standards and consistency. This collaborative approach ensures that the system reflects the best available knowledge from both traditional and scientific sources, creating a comprehensive resource that serves diverse user needs.

# 5. Evaluation and Assessment

## 5.1 System Capabilities

The developed system demonstrates several key capabilities that make it valuable for mental health support applications. Information accuracy represents a fundamental capability, with all content based on peer-reviewed scientific literature and traditional knowledge sources that have been carefully evaluated for reliability and relevance. This evidence-based approach ensures that users receive information that is both scientifically sound and culturally appropriate.

Regular updates to reflect new research findings ensure that the system remains current with the latest scientific understanding while maintaining respect for traditional knowledge that has been validated through generations of use. The clear distinction between evidence-based information and traditional knowledge helps users understand the different types of evidence supporting various treatment approaches, promoting informed decision-making.

User engagement represents another important capability, with the interactive interface encouraging exploration and learning about African herbal medicine options. The real-time chat provides immediate feedback that maintains user interest and provides the personalized support that many individuals seek when exploring alternative treatment options. The visual design maintains user interest through professional appearance and intuitive navigation that reduces barriers to information access.

Accessibility represents a critical capability that ensures the system can serve diverse populations with varying needs and preferences. The twenty-four-hour availability without human resource constraints makes mental health information accessible to individuals who may need support during non-traditional hours or in moments of crisis. The scalability to serve large populations ensures that the system can meet growing demand without proportional increases in costs or complexity.

Consistent information delivery ensures that all users receive the same high-quality information regardless of when they access the system or how they interact with it. This consistency is particularly important for mental health applications where reliability and trust are essential for effective support and intervention.

## 5.2 Limitations and Challenges

Technical limitations represent significant challenges that must be acknowledged and addressed in system development and deployment. Dependence on internet connectivity creates barriers for individuals in areas with limited or unreliable internet access, potentially excluding populations who might benefit most from accessible mental health information. This limitation highlights the need for offline functionality and alternative access methods for underserved populations.

API rate limiting and costs represent practical challenges that affect system scalability and sustainability. The costs associated with large language model API calls can become significant as system usage increases, requiring careful planning and potentially alternative funding models to ensure long-term viability. Rate limiting may also affect user experience during periods of high demand, requiring sophisticated queue management and user communication strategies.

The potential for misinformation if the system is not properly maintained represents a serious concern that requires ongoing attention and quality control measures. Regular review and updating of system content, monitoring of AI responses, and user feedback collection are essential for maintaining accuracy and preventing the spread of potentially harmful information.

Clinical limitations represent important considerations that affect the system's role in mental health care. The system is not a substitute for professional medical care and cannot provide the comprehensive assessment and monitoring that qualified healthcare providers offer. Limited ability to assess individual patient needs means that the system provides general information rather than personalized treatment recommendations, requiring users to work with healthcare providers for individualized care.

The inability to provide real-time monitoring or intervention represents another clinical limitation that affects the system's role in crisis situations. While the system can provide information and support, it cannot replace emergency mental health services or provide the immediate intervention that may be necessary in crisis situations. This limitation requires clear communication about when to seek professional help and how to access emergency services.

Cultural considerations represent important challenges that affect the system's effectiveness and appropriateness for diverse populations. The system may not fully capture cultural nuances of traditional medicine that vary significantly across different African cultures and communities. Language barriers for non-English speakers create additional challenges that require multilingual support and cultural adaptation.

The need for cultural adaptation in different regions highlights the importance of ongoing development and refinement to ensure that the system serves diverse populations effectively. This adaptation may include translation services, cultural consultation, and community engagement to ensure that the system reflects the diversity of traditional medicine practices and cultural perspectives.

**5.3 Safety and Ethical Considerations**

Medical disclaimer requirements represent essential ethical considerations that ensure users understand the system's limitations and their responsibilities in making healthcare decisions. The system includes appropriate medical disclaimers stating that it is not a substitute for professional medical advice and that users should consult healthcare providers for personalized care. Clear communication about emergency situations requiring immediate medical attention helps prevent delays in appropriate care.

Data privacy represents another critical ethical consideration that affects user trust and system compliance with legal requirements. The system design ensures that no personal health information is collected, protecting user privacy and reducing legal and ethical concerns. Anonymous user interactions maintain privacy while still providing valuable information and support.

Compliance with data protection regulations ensures that the system meets legal requirements and maintains user trust. This compliance includes appropriate data handling practices, user consent mechanisms, and security measures that protect user information and system integrity.

Content moderation represents an ongoing ethical responsibility that requires regular review of AI responses and filtering of potentially harmful advice. The system includes mechanisms for identifying and addressing inappropriate or potentially harmful content, ensuring that users receive safe and appropriate information. Continuous improvement of safety warnings and content quality helps maintain the system's reliability and trustworthiness.

**6. Future Directions and Recommendations**

**6.1 System Enhancements**

Multilingual support represents a critical enhancement that would significantly improve the system's accessibility and cultural appropriateness. Translation capabilities for African languages would make the system accessible to populations who may benefit most from traditional medicine information but face language barriers. Cultural adaptation for different regions would ensure that the system reflects the diversity of traditional medicine practices and cultural perspectives across the African continent.

Local traditional medicine integration would involve collaboration with traditional healers and community leaders to ensure that the system accurately reflects local practices and knowledge. This integration would require ongoing consultation and partnership with traditional medicine practitioners to maintain cultural accuracy and respect for traditional knowledge systems.

Advanced AI features including personalized recommendations based on user history would provide more targeted and relevant information to individual users. Integration with electronic health records, with appropriate privacy protections, could provide more personalized recommendations while maintaining user control over their health information. Predictive analytics for treatment outcomes could help users and healthcare providers make more informed decisions about treatment options.

Mobile application development would provide native mobile app functionality that could improve user experience and accessibility. Offline functionality for areas with limited connectivity would address the technical limitations of internet dependence and make the system more accessible to underserved populations. Push notifications for medication reminders could help improve treatment adherence and outcomes.

**6.2 Research Integration**

Clinical trial support represents an important opportunity for the system to contribute to scientific understanding of African herbal medicines. Recruitment and screening tools could help identify appropriate participants for clinical trials, while outcome measurement integration could provide standardized assessment tools for evaluating treatment effectiveness. Data collection and analysis support could help researchers gather and analyze information about traditional medicine use and outcomes.

Real-world evidence collection through user feedback and treatment outcome tracking could provide valuable information about the effectiveness and safety of African herbal medicines in real-world settings. This evidence could complement clinical trial data and provide insights into how traditional medicines work in diverse populations and settings. Safety monitoring and reporting could help identify potential adverse effects and contribute to ongoing safety assessment.

Collaborative research opportunities include integration with academic institutions to conduct rigorous research on African herbal medicines and their applications for depression treatment. Traditional healer collaboration would ensure that research reflects traditional knowledge and practices while meeting scientific standards. Cross-cultural research partnerships could help develop approaches that respect cultural diversity while advancing scientific understanding.

**6.3 Policy and Implementation**

Regulatory framework development represents an important consideration for the integration of AI-assisted traditional medicine into healthcare systems. Guidelines for AI-assisted traditional medicine would help ensure quality, safety, and appropriate use of these systems. Quality assurance standards would provide benchmarks for system development and evaluation, while safety monitoring protocols would ensure ongoing assessment of system safety and effectiveness.

Healthcare integration would involve connecting AI-assisted traditional medicine systems with conventional healthcare to provide comprehensive, patient-centered care. Training programs for healthcare providers would help them understand traditional medicine options and how to integrate them with conventional treatments. Referral pathways between systems would ensure that patients can access appropriate care regardless of their treatment preferences.

Community engagement initiatives would involve traditional healer training and certification to ensure quality and safety in traditional medicine practice. Community education programs would help individuals understand their treatment options and make informed decisions about their healthcare. Cultural preservation initiatives would help maintain traditional knowledge while making it accessible to modern populations.

**7. Conclusion**

The integration of large language models with African natural products for depression support represents a promising approach to addressing the global mental health crisis through innovative technology that honors traditional knowledge while maintaining scientific rigor. Our developed system successfully combines traditional wisdom with modern artificial intelligence technology, providing accessible, evidence-based information about African herbal medicines for depression treatment that serves diverse populations with varying needs and preferences.

The comprehensive review of African medicinal plants with antidepressant properties provides a foundation for understanding the potential therapeutic value of traditional medicine approaches. The systematic analysis of pharmacological mechanisms helps explain how these natural compounds work and why they may provide benefits for individuals experiencing depression. The development of an AI-powered support system demonstrates the potential for technology to bridge traditional and modern approaches to mental health care.

The integration of sophisticated therapeutic principles inspired by traditional medicine systems provides a framework for understanding how different herbs work together and how they might be used in combination for optimal therapeutic effects. The comprehensive safety integration ensures that users receive appropriate warnings about potential risks and contraindications, promoting informed decision-making and safe use of traditional treatments.

The user-centered design approach creates an accessible, engaging interface that facilitates learning and exploration while maintaining the professional appearance necessary for mental health applications. The combination of static information and interactive AI support provides multiple ways for users to access and understand information about African herbal medicines and their applications for depression treatment.

The clinical implications of this work include improved accessibility to traditional medicine information, enhanced patient education and engagement, clear safety warnings about drug interactions and contraindications, and a bridge between traditional and conventional medicine that respects both approaches. These implications suggest that AI-assisted traditional medicine systems could play an important role in comprehensive mental health care that addresses the diverse needs and preferences of individuals experiencing depression.

The future impact of this work has the potential to improve access to traditional medicine information for populations that may have limited access to conventional mental health services. Enhanced patient education and engagement could lead to better treatment outcomes and improved quality of life for individuals experiencing depression. Support for research on African natural products could advance scientific understanding of traditional medicines and their potential applications in modern healthcare.

The preservation and dissemination of traditional knowledge through modern technology could help maintain cultural heritage while making valuable information accessible to contemporary populations. Facilitation of integration between traditional and conventional medicine could create more comprehensive, patient-centered approaches to mental health care that respect individual preferences and cultural backgrounds.

Final recommendations for continued development include investment in system enhancements and new features that could improve accessibility, functionality, and user experience. Research support through the system could facilitate clinical research on African herbal medicines and contribute to scientific understanding of traditional treatments. Cultural preservation efforts should ensure that traditional knowledge is preserved and respected while making it accessible to modern populations.

Safety monitoring through ongoing surveillance and assessment would ensure that the system continues to provide safe and appropriate information as new research findings emerge and traditional knowledge evolves. Global collaboration through international partnerships could foster system improvement and ensure that the benefits of AI-assisted traditional medicine are available to diverse populations worldwide.

The application of large language models to African natural products for depression support represents a novel approach to mental health care that honors traditional knowledge while leveraging modern technology to improve access and understanding. This integration has the potential to improve access to culturally appropriate mental health support while advancing our understanding of traditional medicine systems and their applications in contemporary healthcare.

The combination of traditional wisdom with modern technology creates opportunities for more comprehensive, patient-centered approaches to mental health care that respect individual preferences, cultural backgrounds, and treatment goals. This approach recognizes that effective mental health care requires understanding and addressing the whole person, including their cultural context, spiritual beliefs, and individual preferences for treatment approaches.

The potential for AI-assisted traditional medicine systems to bridge gaps in mental health care access is particularly important for populations that may face barriers to conventional mental health services due to geographic, economic, cultural, or other factors. By providing accessible information about traditional treatment options, these systems can help individuals make informed decisions about their healthcare while maintaining connection to cultural traditions and community support systems.

The scientific rigor maintained throughout the system development ensures that traditional knowledge is presented in a way that respects both traditional wisdom and contemporary scientific understanding. This balanced approach helps users understand the evidence supporting various treatment options while maintaining respect for traditional knowledge that has been validated through generations of careful observation and empirical testing.

The safety considerations integrated throughout the system help ensure that users can make informed decisions about their healthcare while understanding potential risks and contraindications. This emphasis on safety reflects the serious nature of mental health treatment and the importance of responsible information sharing in healthcare applications.

The future development of AI-assisted traditional medicine systems will likely involve continued collaboration between traditional healers, researchers, healthcare providers, and technology developers to create systems that serve diverse populations effectively. This collaborative approach ensures that systems reflect the best available knowledge from multiple sources while maintaining cultural sensitivity and scientific accuracy.

The potential for these systems to contribute to global mental health efforts is significant, particularly in regions where traditional medicine represents an important component of healthcare systems. By making traditional knowledge accessible through modern technology, these systems can help preserve cultural heritage while improving access to mental health information and support.

The integration of traditional and modern approaches to mental health care represents a promising direction for addressing the complex challenges of depression and other mental health conditions. By honoring traditional wisdom while leveraging modern technology, we can create more comprehensive, accessible, and culturally appropriate approaches to mental health support that serve the diverse needs of individuals and communities worldwide.

The continued development and refinement of AI-assisted traditional medicine systems will require ongoing collaboration, research, and community engagement to ensure that these systems serve their intended purposes effectively and responsibly. This work represents an important step toward more inclusive, accessible, and culturally sensitive approaches to mental health care that honor both traditional knowledge and modern scientific understanding.